\documentclass[conference]{IEEEtran}

\usepackage{cite}
\usepackage{amsmath,amssymb,amsfonts}
\usepackage{algorithmic}
\usepackage{graphicx}
\usepackage{textcomp}
\usepackage{subfigure}
\usepackage{xcolor}
\usepackage{tabularx}
\def\BibTeX{{\rm B\kern-.05em{\sc i\kern-.025em b}\kern-.08em
    T\kern-.1667em\lower.7ex\hbox{E}\kern-.125emX}}

\begin{document}

\title{Enhanced Monocular Visual Odometry with AR Poses and Integrated INS-GPS for Robust Localization in Urban Environments}
\author{\IEEEauthorblockN{Ankit Shaw}
\IEEEauthorblockA{\textit{College of Engineering} \\
\textit{University of Washington}\\
Seattle, WA, USA \\
ankit25@uw.edu}
}

\maketitle

\begin{abstract}
    This paper introduces a cost-effective localization system combining monocular visual odometry (VO), augmented reality (AR) poses, and integrated INS-GPS data. I address monocular VO’s scale factor issues using AR poses and enhance accuracy with INS and GPS data, filtered through an Extended Kalman Filter (EKF). My approach, tested using manually annotated trajectories from Google Street View, achieves an RMSE of 1.529 meters over a 1 km track. Future work will focus on real-time mobile implementation and further integration of visual-inertial odometry for robust localization. This method offers lane-level accuracy with minimal hardware, making advanced navigation more accessible.
\end{abstract}

\section{Introduction}
In recent years, many cars have released with SAE Level 2 autonomous driving features. The price tag for a car with these new features is still out of reach of many people, including those who may need them most. One of the major contributors to the price of these vehicles is the cost of their sensors. Currently, a full suite of sensors for a level 2 autonomous vehicle can cost anywhere from \$8,000-10,000\cite{sensorcost}. As I begin to progress towards higher levels of autonomy, the suite of sensors needed will only increase in complexity, and in turn, price. Additionally, it has been shown that traditionally GPS is not accurate enough to be used as a localization method for autonomous driving and this inaccuracy only increases exponentially in an urban environment. With my approach I aim to create a level of accuracy that is comparable to that of much more expensive sensors at a fraction of the cost. 

In this paper I proposed a way to exploit this ‘urban jungle’ and improve greatly upon the accuracy of a traditional GPS using sensors found on a single smartphone. My method is based around using Google Street View’s panoramic images and comparing them to those being taken on board with my monocular smartphone camera. I chose to use Street View because it contains data from a large portion of the world and this data gives me both a high resolution image with accurate associated location metadata. Additionally, anyone around the world with an internet connection can access Street View. I also log measurements from the IMU and Apple's ARKit Framework simultaneously with the images recorded from my smartphone. 

My approach involves extracting features from my recorded frames, as well as from the panoramas taken from Google Street View. Once I have these features, I calculate the relative pose between my frame and the surrounding panoramas. I then use the poses in tandem with the coordinates of each panorama provided in the metadata to triangulate my estimated position. Additionally, I aim to use the IMU measurements alongside my image data to compute my Visual-Inertial Odometry module and combine it with my estimated location in order to reduce noise.

\section{Related Work}
Previous work exists on visual global localization, including work specifically on the Google Street View dataset. Agarwal et al. \cite{metric} used optical-flow to create tracks of 3D points and performs a semi-global bundle adjustment on these tracks. More recently, Yu et al. \cite{msthesis} took a more limited approach by only performing a local bundle adjustment on the current set of images, not taking advantage of any odometry data. This approach showed centimeter-level accuracy, however it also made use of depth data that I found unreliable in many areas, and two roof-mounted sideways cameras that faces street facades. My approach builds off of this previous work from \cite{metric} and \cite{msthesis} but uses only a single front-facing phone camera and combines additional odometry data directly from the phone.

\section{Method}

\begin{figure}[h]
\centerline{\includegraphics[width=0.8\linewidth]{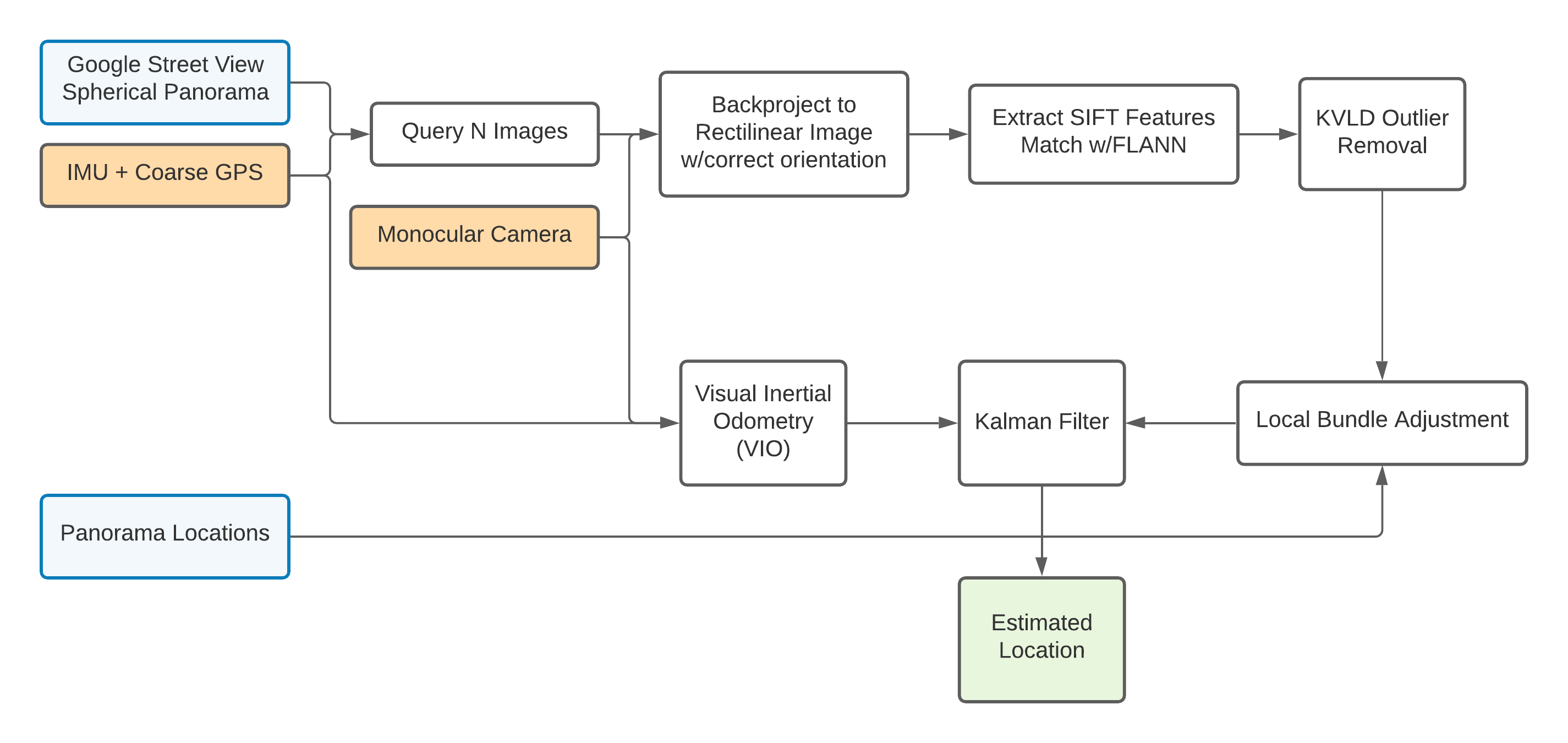}}
\caption{Diagram illustrating my system pipeline}
\label{system}
\end{figure}

In this section, I describe my pipeline for global localization. The pipeline was implemented in an offline manner, but at each timestep I only look at current inputs and the estimated vehicle pose, discarding all other past inputs. Although I do perform bundle adjustment, my approach does not involve SLAM in its typical formulation since I do not perform mapping. Instead, I seek to only compute a global localization through feature matching and local bundle adjustment in combination with Visual-Inertial Odometry. The diagram in Figure \ref{system} illustrates the important steps of my pipeline.

\subsection{Google Street View Data}

Google Street View is a platform that contains millions of panoramas and depth maps, collected from vehicles driving throughout the world. Each panorama is a 16384x8192 equirectangular image with a 360° horizontal and 180° vertical field-of-view (FoV), and each depth map is a 512x256 image with the same horizontal and vertical FoV. The panoramas are tagged with precise accurate latitude and longitude coordinates which is calculated by first fusing sensor data from GPS, wheel encoders, and an IMU, and further optimizing the trajectory with known road network and intersection maps \cite{gsv}. 

\begin{figure}[h]
\centerline{\includegraphics[width=0.8\linewidth]{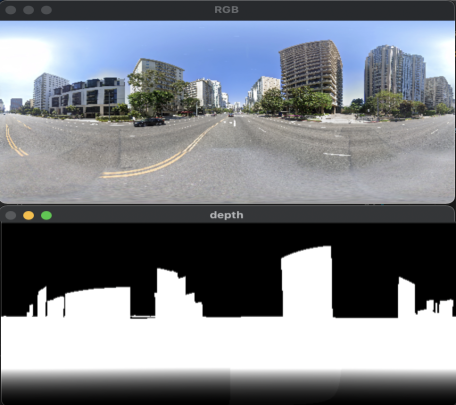}}
\caption{Google Street View Panorama (RGB and Depth)}
\label{streetview}
\end{figure}

The depth map captures data from laser scans or using optical flow methods, and stores only the dominant scene features such as buildings and roads \cite{gsv}. However, as seen in Figure \ref{streetview}, in dense urban environments where I need complete representation of static scene elements, the depth map is unreliable in generating that data.

To perform feature extraction and matching on the panoramas, I must generate rectilinear images from the equirectangular panoramas. Google provides an API to pull rectilinear images at specified panorama locations given the heading and pitch angles. Instead, I store a database of panoramas I intend to query as equirectangular projections of the full 360° field-of-view, and extract the rectilinear regions of the panorama with the precise vehicle heading, making my method more robust and fast. I create a “virtual camera” and perform a back-projection technique to project the panorama into a standard rectilinear view. Provided a heading and field of view, I can compute the corresponding pixel coordinates on the rectilinear image from the panorama. With the center elevation and azimuthal angles of the image being $ (\theta_0, \phi_0) $ and the angular distance from rectilinear pixel coordinates $ (x, y) $ to panorama coordinates $ (\theta, \phi) $ defined as $c$, the transformation equations described in \eqref{backprojection} are used to generate rectilinear images with a defined focal length and FoV.

\begin{equation}
\begin{aligned}
\begin{split}
& x=\frac{cos(\theta)sin(\phi-\phi_0)}{cos(c)}
\\
& y=\frac{cos(\theta_0)sin(\theta) - sin(\theta_1)cos(\theta)cos(\phi-\phi_0)}{cos(c)}
\\
& cos(c)=sin(\theta_0)sin(\theta) + cos(\theta_0)cos(\theta)cos(\phi-\phi_0)
\end{split}
\end{aligned}
\label{backprojection}
\end{equation}

This virtual camera has the same camera intrinsics as my vehicle camera, maintaining the same focal length, horizontal and vertical field-of-view, and resolution, although I assume the principal point of the camera is the image center. Consistency in the camera intrinsics will simplify future computations for 3D point estimation and pose estimation knowing that the focal lengths are the same.

\subsection{Feature Matching}
\begin{figure}[h]
\centerline{\includegraphics[width=0.8\linewidth]{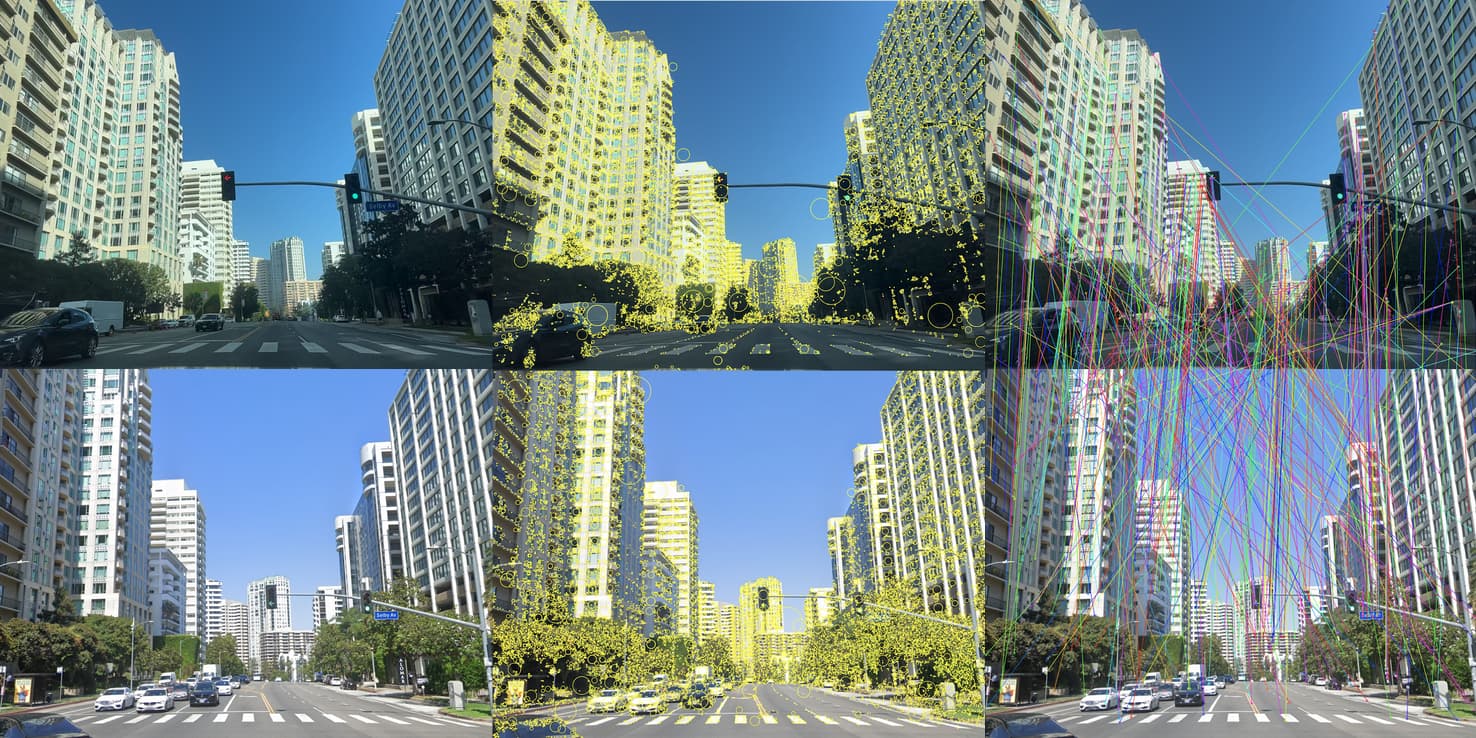}}
\caption{From left to right: captured frame and nearby panorama image, camera frame and panorama SIFT key points, camera frame and panorama FLANN matches.}
\label{sift_flann}
\end{figure}

For each frame in my video stream, I find nearby Street View panoramas using a KD-tree. For my image frame and one of these nearby panoramas, I use SIFT to extract key points in both images, shown in \ref{sift_flann} \cite{sift}. With these SIFT key points, I then use FLANN based matching to find matching key points between my image and the panorama, similarly shown in \ref{sift_flann} \cite{flann}. From this example set of matches, I notice no clear set of matching features between the two images.  Clearly, a large portion of the computed matches in this example are incorrect and as a result should be discarded. These incorrect matches, referred to onwards as outliers, negatively affects my ability to find the essential matrix that describes the epipolar geometry between my captured image and the nearby panorama, which leads to inaccurate localization.

\begin{figure}[h]
\centerline{\includegraphics[width=0.8\linewidth]{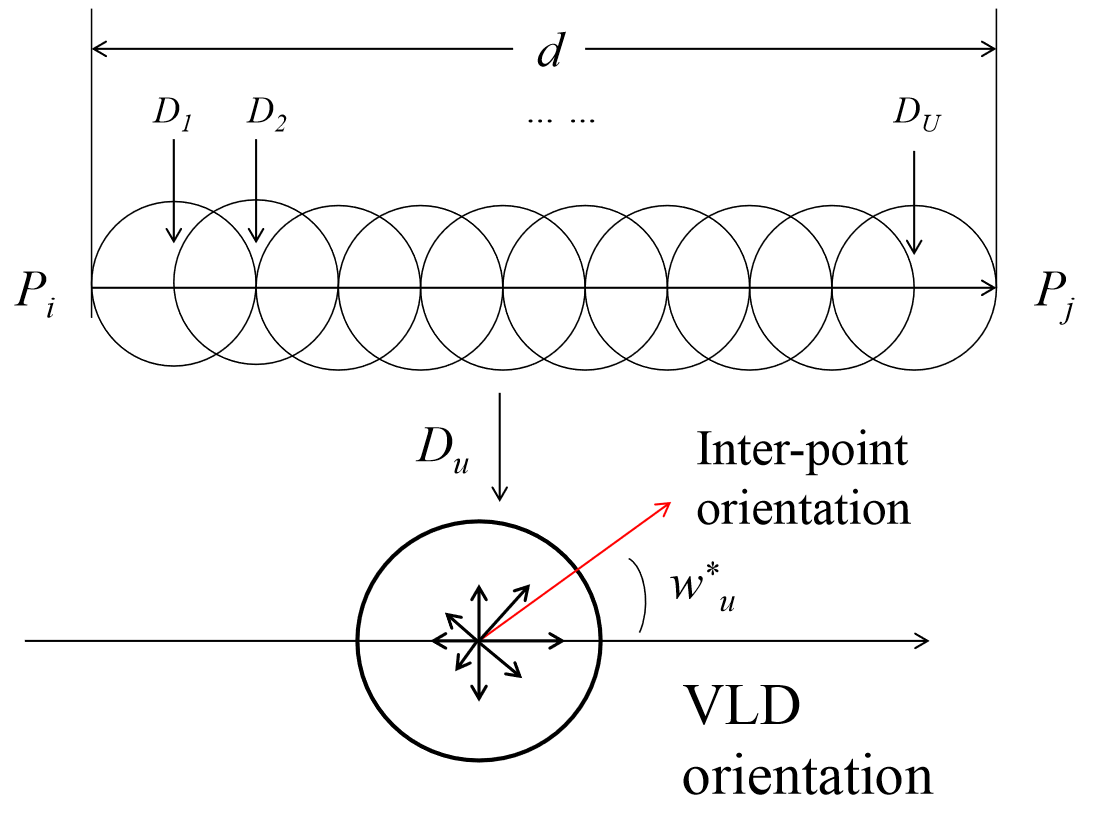}}
\caption{Visualisation of a Virtual Line Descriptor \cite{kvld}}
\label{vld}
\end{figure}

To effectively address the outlier issue, I apply a filtering technique to my FLANN matches using second order descriptors known as Virtual Line Descriptors (VLD) \cite{kvld}. To define a VLD, for each pair of key points in an image I compute a number of separate SIFT-like descriptors for the region of the image between the two points, shown in \ref{vld}. Together, this collection of descriptors act as my second order descriptor, or VLD. With this definition, I can then also compute the difference between two VLDs as the pairwise difference between corresponding disks in their gradient histograms as well as main orientations \cite{kvld}.

Now, I consider a pair of matches in my image and the panorama, giving me two points, or one VLD, in each of the two images. Then, I find the difference between these two VLDs, and if this difference is below a maximum threshold I consider this pair of matches to be consistent \cite{kvld}.

The K-VLD filtering algorithm uses this idea of consistent matches, and for any primary match between my image and the panorama, I look one by one at all the nearby matches, and for each of these, check if the pair of matches is consistent \cite{kvld}. If there are at least K consistent matches for my primary match, it satisfies the K-VLD consistent constraint and is kept as a reliable match \cite{kvld}. This process is repeated for all matches between the two images until I have only K-VLD consistent matches remaining \cite{kvld}.

\begin{figure}[h]
\centerline{\includegraphics[width=0.8\linewidth]{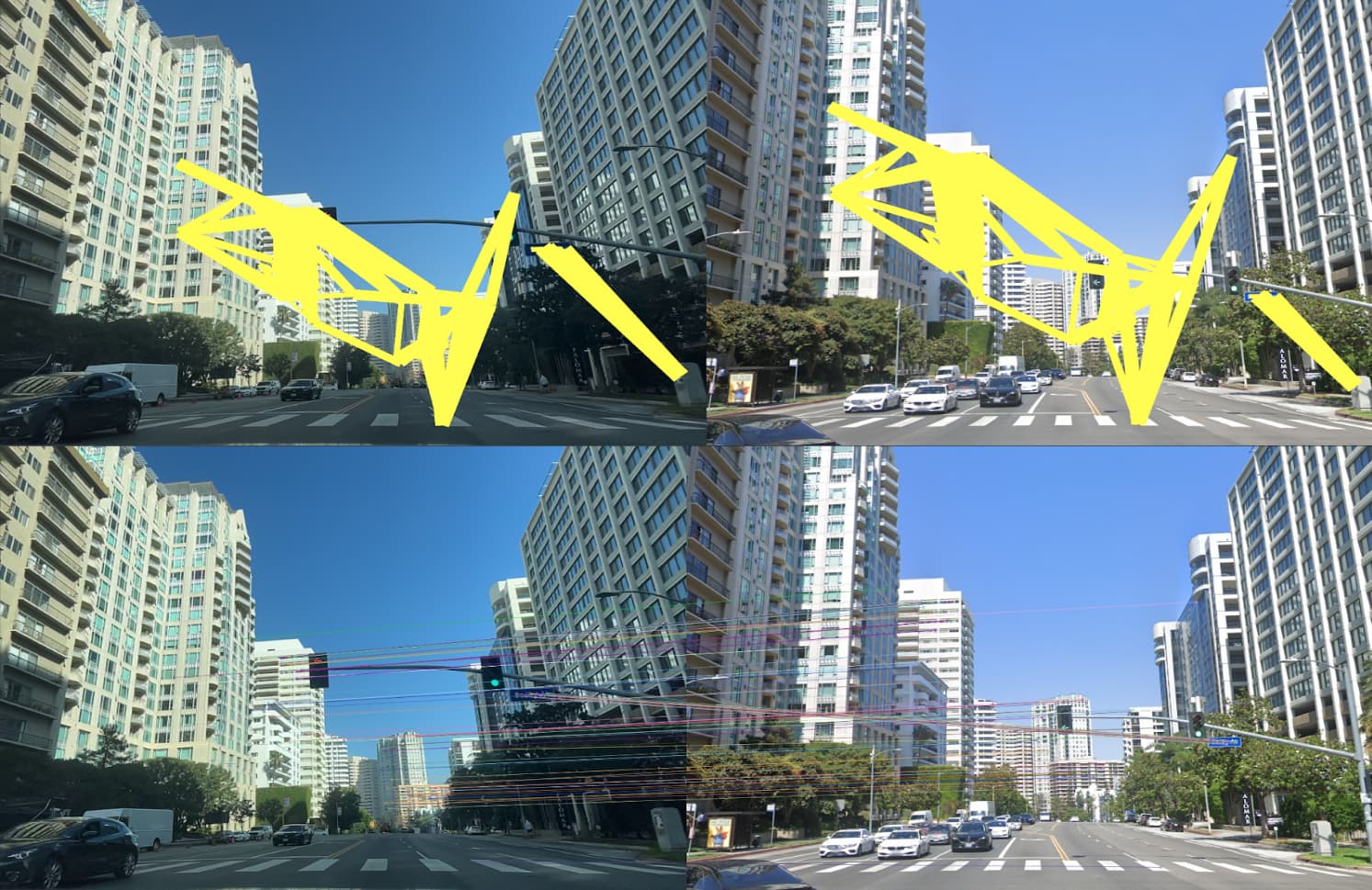}}
\caption{Top: Camera frame and panorama with consistent VLDs highlighted. Bottom: K-VLD filtered matches }
\label{kvld_filtered}
\end{figure}

I return to my previously computed FLANN matches in \ref{sift_flann} with all of the outliers which I need to remove. Using the K-VLD algorithm, I can find all of the consistent VLDs, shown in \ref{kvld_filtered} where these VLDs are highlighted. I then keep all matches from these consistent VLDs, leaving a much smaller set of matches, but now with almost all outliers removed, shown in \ref{kvld_filtered}.

The main benefit of K-VLD filtering for my application is its property of removing ambiguous matches \cite{kvld}. These types of matches are common in my dataset, which features large numbers of ambiguous features such as building windows. Additionally, this method removes false matches near epipolar lines, which again is well suited for my dataset with the majority of detected features being located along or near epipolar lines \cite{kvld}. I perform SIFT feature extraction, FLANN matching, and K-VLD filtering using a fork of openMVG \cite{mvg}.

\subsection{Visual Pose Estimation}
In order to estimate the pose of the vehicle, I tested two similar but distinct strategies, both based on a local bundle adjustment. The first method relies upon known fixed panorama locations and uses a limited local bundle adjustment to estimate 3D points common between all panorama images. Then, using features common with the phone camera, I use PnP to estimate the phone camera relative to a reference panorama. The second method involves a standard local bundle adjustment procedure where I allow the known panorama locations as well as the vehicle pose to be adjusted subject to some constraints and use the final vehicle pose from this computation. I chose these two approaches because I sought to analyze whether allowing for more flexibility during bundle adjustment w.r.t camera poses would produce more accurate localization.

\subsection{Fixed-pose Local Bundle Adjustment}

\begin{figure*}[h]
\centering
\subfigure[]{\includegraphics[width=.32\textwidth]{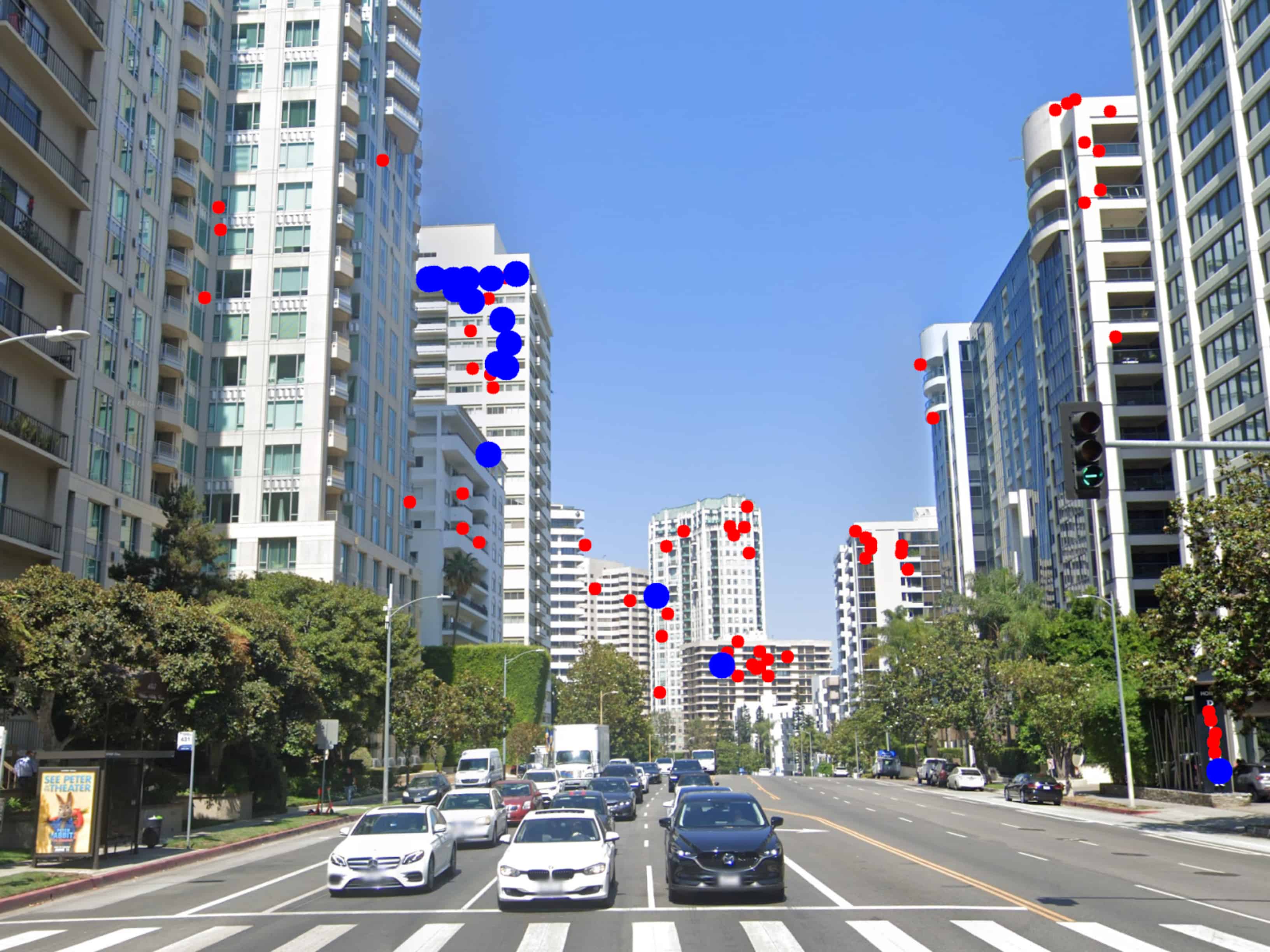} }
\subfigure[]{\includegraphics[width=.32\textwidth]{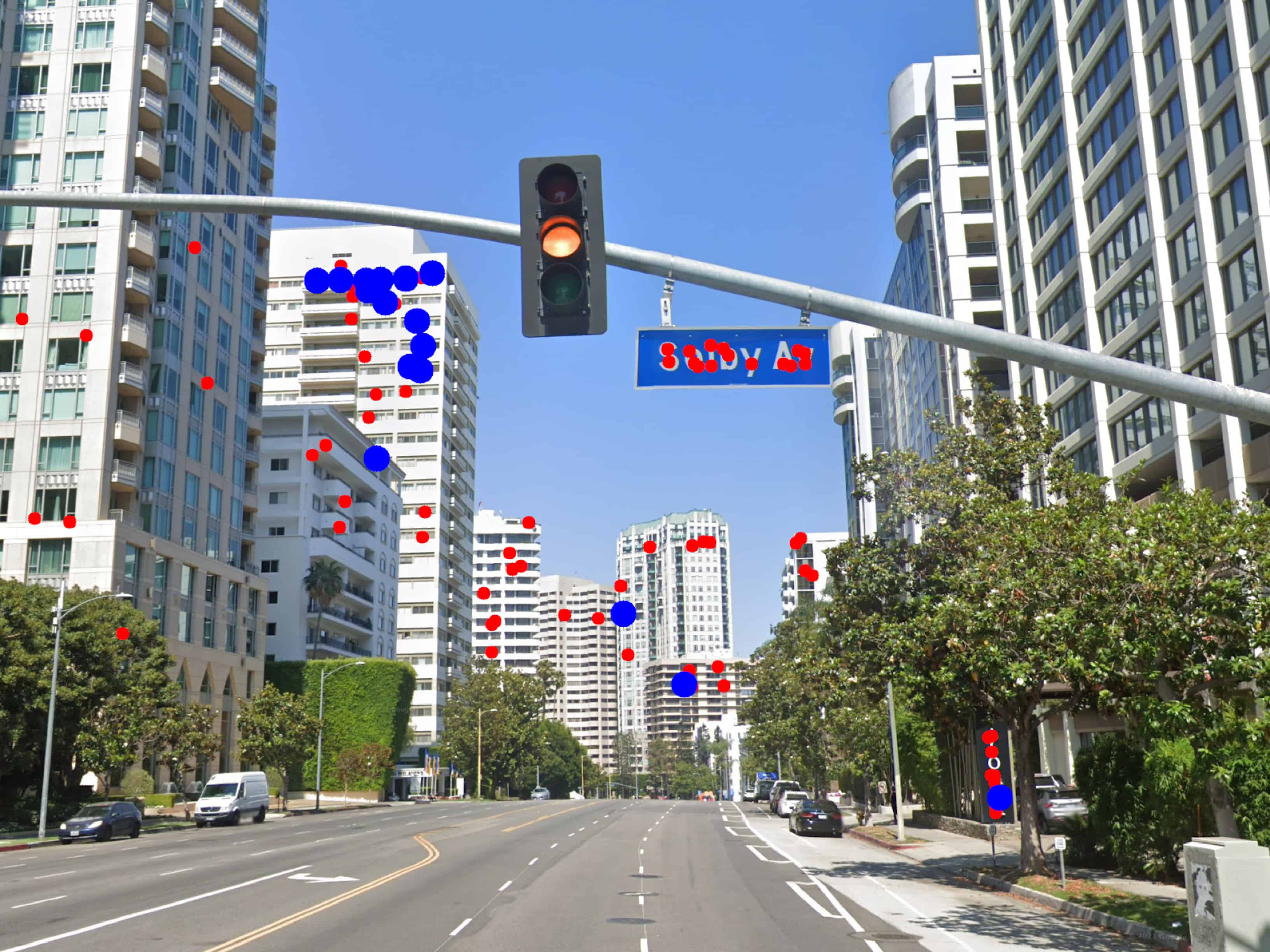} }
\subfigure[]{\includegraphics[width=.32\textwidth]{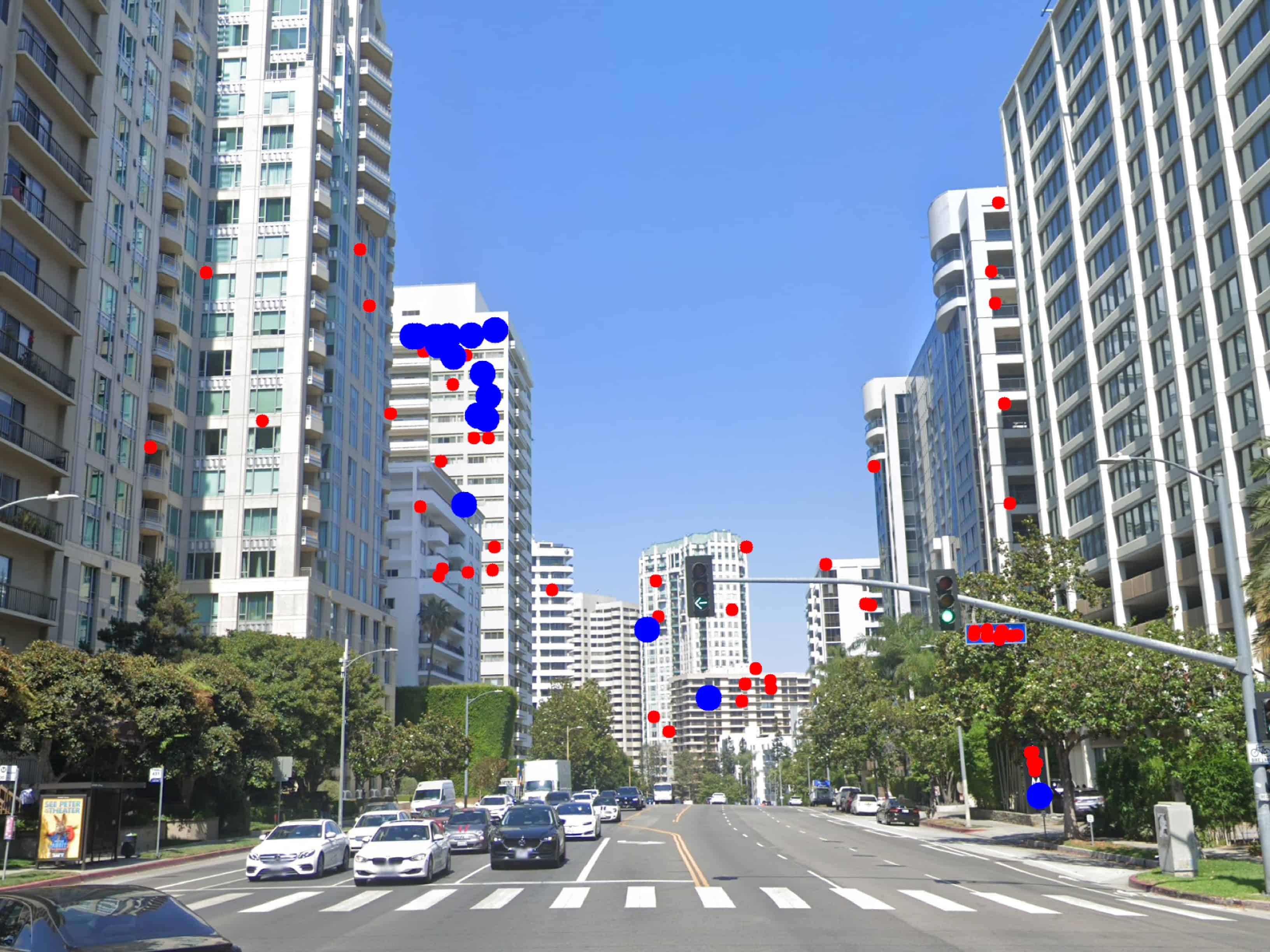} }
\caption{Features present in all panorama views. Red key points indicate features unique to that frame and blue key points indicate features found in all frames.}
\label{fig:features2}
\end{figure*}

With a set of panoramas  $P$ and their corresponding features $F$ and matches $M$ with the vehicle frame $V$, I will take the subset of features $$ F^* = \{F_P \cap F_V\} $$ that are present in all panoramas $P$ and vehicle frame $V$. This will give me a set $M^*$ containing pairs of features, where for each feature in $F_V$ I will have a corresponding feature $F_P$ in each of the panoramas, as seen in Figure \ref{fig:features1}\ref{fig:features2}. I can then perform a nonlinear optimization using SciPy to estimate the 3D points for each feature $F_V$ relative to the selected panorama by minimizing the following loss function. Unlike a typical bundle adjustment where I allow the camera poses to be adjusted, I fix these poses and only adjust the 3D point estimations while minimizing the error defined in \eqref{3d_point_estimation}.

\begin{equation}
F(x,y)=\sum_{ij} e_{ij}(x,y)
\label{3d_point_estimation}
\end{equation}
where: \\
\begin{tabularx}{\linewidth}{lX}
$x$ &= A vector of 6DOF Street View camera poses\\
$y$ &= A vector of 3D points associated with each feature\\
$e_{ij}$ &= Error function between predicted 3D points $z_{ij}(x,y)$ and measured point $\hat{z}_{ij}=(\theta_{ij}, \phi_{ij})$\\
$z_{ij}$ &= The elevation and azimuthal angles from camera pose x to 3D feature point in the 2D camera frame\\
\end{tabularx}
\\\\

\begin{figure}[h]
\centerline{\includegraphics[width=0.8\linewidth]{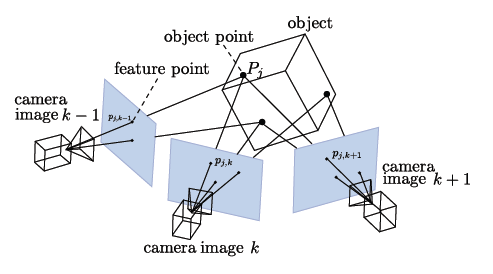}}
\caption{Multi-view 3D Point Estimation. Object points are estimated knowing that a corresponding feature point is in each camera view. Courtesy of \cite{9}}
\label{fig:3d_points_mvg}
\end{figure}

With the image coordinates and 3D global coordinates of $F_V$, I can perform a Perspective-n-Point pose estimation to best estimate the pose of the camera relative to the 3D points to best minimize the reprojection error. The overall problem of 3D point estimation can be seen in Figure \ref{fig:3d_points_mvg}, where I estimate the positions of the 3D object points when each of the corresponding feature points in each camera view is known.

\subsection{Standard Local Bundle Adjustment}

As before, I have a set of panoramas  $P$ and their corresponding features $F$ and matches $M$ with the vehicle frame $V$. Instead of taking the intersection of features $$ F^* = \{F_P \cap F_V\} $$ that are present in all panoramas $P$, I instead require only that a feature is present in at least $n$ panoramas. This will give me a set $M^*$ containing pairs of features, where for each feature in $F_V$ I will have a corresponding feature $F_P$ in at least $n$ panoramas.

To get the best results from the local bundle adjustment, I need to initialize the solver with an estimate of all $|M^*|$ 3D points. Since each of my $|M^*|$ 3D points is present in at least $n$ panoramas, there is not one correct initialization. Instead, I use the the 8-pt algorithm for each $(f_v, f_p), f_v \in F_V, f_p \in F_p$ pair to approximate the essential matrix. I then recover the relative pose transformation between the cameras and perform 3D reconstruction by triangulation. I then retrieve only the 3D points that are common to at least $n$ images.

Thus I have at least $n$ estimates of any given 3D point through my approximate reconstruction method. However, these points are in the reference frame of the respective panoramas so I must transform these coordinates into a global frame which I define as the reference frame of the panorama with highest number of feature matches. Given at least $n$ estimates of any given 3D point I pick the point with the median L2 norm as a simple form of outlier rejection.

\begin{figure}[h]
\centerline{\includegraphics[width=0.8\linewidth]{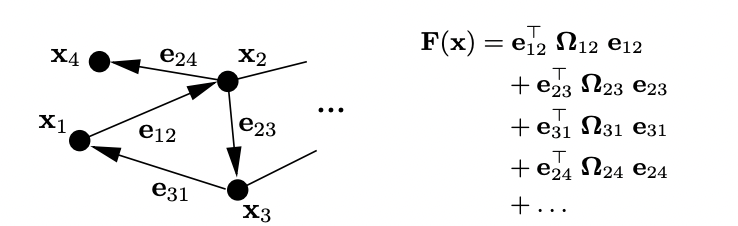}}
\caption{Relationship between objective function and factor graph courtesy of \cite{g2o}}
\label{factor}
\end{figure}

I chose to compare two common nonlinear solvers used to for graph optimization, g2o \cite{g2o} and Ceres \cite{ceres}. A graph is a natural representation of the bundle adjustment problem since it allows me to relate quantities I wish to estimate such as camera poses and 3D points, with my measurements as edges that relate the quantities. Specifically, I construct a graph such that the edges represent the projection of landmark 3D points onto my camera frames. A visual representation of how the factor graph represents an objective function is seen in figure \ref{factor} courtesy of \cite{g2o}.

I create $n$ vertices in my factor graph $G$ corresponding to my $n$ panorama locations and, for Ceres, I fix the allowable translation and rotation of these poses to some upper and lower bound. With more data on the accuracy of these panorama poses, future work may incorporate a Gaussian prior term in the error function that represents my certainty of the panorama poses and penalizes significant adjustments. This would better model my assumption of the system but I were unable to incorporate it into the Ceres solver through the Python bindings used for testing.

Next, I add an additional vertex corresponding to my phone camera pose which I initialize from my Kalman Filter or, for the first frame, my GPS coordinates. This vertex is entirely unconstrained.

For each of my $|M^*|$ 3D points, I add a vertex to my factor graph $G$ with the estimated position from my 3D reconstruction and attach at least $n + 1$ edges to this point. At least $n$ edges are between panorama locations and the point with an additional edge connecting the smartphone camera frame. These edges contain the observed $(x,y)$ pixel that corresponds to the 3D point as well as the uncertainty of the measurement. The objective that I seek to minimize is the reprojection error of these 3D points onto the least $n + 1$ cameras in their respective locations, and I do so by adjusting the positions of the 3D points and the 6-DoF poses of the cameras.

As suggested in \cite{msthesis} I use the Tukey Biweight function as in \eqref{tukey} with $c = 3$ to improve convergence of my optimization problem in cases where outliers remain past the K-VLD step. I replace the squared error I are trying to minimize with the Tukey Biweight function and use g2o and Ceres to solve the graph optimization problem.

\begin{equation}
\begin{aligned} \ell (r) = \begin{cases} \frac{c^2}{6} \left(1 - \left[ 1 - \left( \frac{r}{c}\right)^2 \right]^3 \right) &\text{if } |r| \leq c, \\ \frac{c^2}{6} &\text{otherwise}. \end{cases} \end{aligned} \label{tukey}
\end{equation}

\subsection{Sensor Fusion}
my estimate of the vehicle’s position using Google Street View might suffer from various points of failure such as occlusion of static features, lighting and weather condition and inaccurate street view data. The dependability of autonomous systems of one just one such position estimate can result in serious consequences. I address this problem by implementing a multi-sensor fusion algorithm. The recent trend in high quality and low cost smartphones has led to multiple sensors being incorporated into a small form factor. I exploit these sensors to derive a better estimate of the vehicle's state. I explored two such methods:
\subsubsection{Visual Odometry and AR Poses}
Visual odometry is estimating the motion of the vehicle using visual cues from the cameras attached to it. I deployed a featured based monocular odometry. my  pipeline for visual odometry is as shown in Fig. 5. I use the FAST corner detection algorithm \cite{1}\cite{2} to detect features in each frame It due to it’s fast computational efficiency. Each of these corner pixels are tracked using the optical flow bases KLT tracker \cite{3} in the successive frame It+1. By selecting points using RANSAC from pairs of corresponding points in successive frames I estimate the essential matrix E as shown in Eqn. 4 using the Nister five-point algorithm\cite{4}. Furthermore, decomposing the essential matrix using SVD and exploiting rotation matrix constraints gives the rotation and translation matrix between the two consecutive frames as shown in Eqn. 5 and 6.
\begin{equation}
y_1^T E y_2 = 0
\end{equation}
\begin{equation}
E = U \Sigma V^T
\end{equation}
\begin{equation}
  [t] = V W \Sigma V^T ,\; R = U W^{-1}V^T
\end{equation}

The net pose with respect to the first frame is given by 
\begin{equation}
R_{pos} = R R_{pos}, \; T_{pos} = t_{pos} +\mu t R_{pos}
\end{equation}

Translation in Monocular visual odometry suffers from scale factor issues that need to be estimated using external sensors. I use AR poses as obtained from the Augmented Reality Toolkit present in iPhone. The relative scale is evaluated by calculating the Frobenius norm between the transnational vectors of the two frames.

\begin{figure}[h]
\centerline{\includegraphics[scale=0.3]{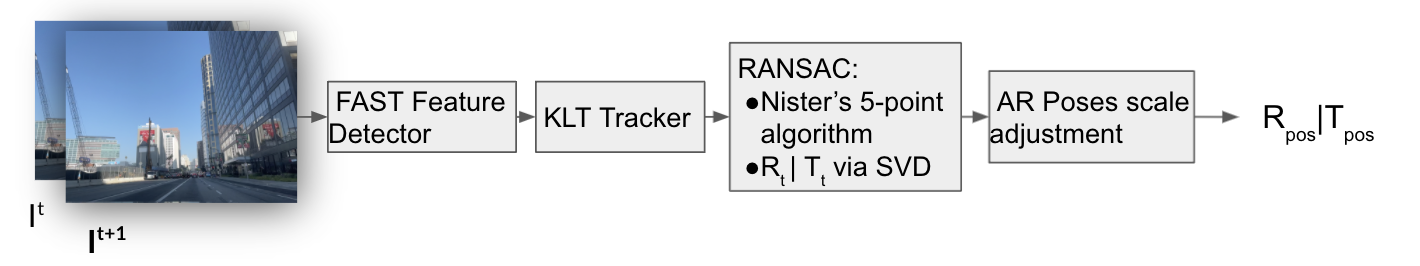}}
\caption{Visual odometry pipeline with AR pose scale}
\label{3d_points}
\end{figure}

\subsubsection{INS and Estimated GPS}
Initial Navigation system (INS) is a state estimation system that uses data from a IMU to continuously calculate the attitude of the system by dead reckoning (Position, Velocity and Orientation). IMU consists of a accelerometer, gyroscope and magnetometer that measures the linear acceleration, angular velocity and magnetic field of the Vehicle respectively for all 3 directions. I used a strapdown INS that integrates angular rate in attitude estimation and uses the attitude data to transform the IMU acceleration into latitude, longitude and heading. A detailed explanation and equations used for INS are given here \cite{6}\cite{7}. The output trajectory is prone to gyroscope and accelerometer bias and random noise that gets added during the dead reckoning process to cause a drift from the actual trajectory. I used an Extended Kalman Filter that inputs high-frequency IMU data and predicts the state of the system based on the INS one time stamp ahead. It also updates the error covariance of the EKF filter. For each low-frequency GPS data I update filter state and error covariance by computing the Kalman gain. The state for the system is given as: $ x= \begin{bmatrix}
q_0 & q_1 & q_2 & q_3 & pos_N & pos_E & V_N & V_E
\end{bmatrix}^T$
And the state space equation as:\\
        $x_{t|t-1} = f(\hat{x}_{k-1|k-1},u_k)=$

Where:\\
$u_k$ = is controlled by accelerometer and gyroscope data that has been converted to delta velocity and delta angle through trapezoidal integration\\
$q_i$ : Parts of the orientation quaternion\\
$pos_i$ : Position in local NED frame\\
$\Delta \theta bias_i$ : Bias in the integrated gyroscope reading\\
$\Delta  vbias_i$ : Bias in the integrated accelerometer reading \\
$\Delta \theta_i$ : integrated gyroscope reading\\
$\Delta  v_i$ : integrated accelerometer reading \\
Please refer to \cite{9} for more information on Extended Kalman Filter and \cite{10} for implementation details.  

\section{Testing Methodology}
Since I aim to show that my system performs well with the vast Google Street View dataset, I cannot test with standard autonomous driving datasets such as KITTI. This complicates direct comparisons to alternate methods and requires me to source my own ground truth. A common source of ground truth for global localization tasks is the use of an RTK-GPS system which are capable of centimeter-level accuracy in ideal conditions. However, I were not able to acquire an RTK system and instead rely on a manually annotated trajectory on top of a satellite map. For my data collection, I drove in a single lane with a front-facing camera and later approximated the camera trajectory with a piecewise linear segments. While it is difficult to ascertain the exact accuracy of this method, I expect the ground-truth trajectory to be within 1m of the true trajectory at any given point due to the width of the lane and position of the camera.

In order to evaluate my method of localization, I compare each of the estimated global locations represented by a latitude and longitude using the WGS-84 projection to my ground truth trajectory. I find the minimum distance from that point to the piecewise linear trajectory and then calculate the RMSE of a sequence of estimated points. my estimated trajectory is formulated as a sequence of points, $T_{est}=\{(x_0, y_0), \dots, (x_n, y_n)\}$ and my ground truth trajectory is similarly: $T_{truth}=\{(x_0, y_0), \dots, (x_m, y_m)\}$. 

I define my error function for any $p\in T_{est}$ in \eqref{err} 
\begin{equation}
\operatorname{err}(p)=\|\underset{i,j\;}{\mathrm{argmin}}(\operatorname{dist}(i,j,p))-p\|_2 
\label{err}
\end{equation}

where $0 \leq i < j \leq m$, $i + 1 = j$, and $\operatorname{dist}(i,j,p))$ represents the minimum distance from $p$ to the line defined by the points $i, j$. I finally define the RMSE in \eqref{err1}:

\begin{equation}
RMSE(T_{est})=\sqrt{\frac{1}{|T_{est}|}\sum\limits_{p \in T_{est}} \operatorname{err}(p)}
\label{err1}
\end{equation}

This method does not capture time information and thus would not tell me whether my estimated points are ahead or behind the ground truth trajectory in time. Similarly, since I do not have time information for my ground truth, I cannot plot the error over time and instead choose to plot over frames. This means my RMSE is dependent on factors such as vehicle speed, unexpected stops, etc. A possible approach might be to use the coarse GPS data to create a plot of error versus position but I chose to not implement this since it adds additional sources of error into my measurement that I cannot account.

Another unexpected source of error came from the panorama poses themselves, as provided by the Google Street View API. Despite the high-accuracy expected given the laser-range data and RTK-GPS used by Google Street View cars \cite{gsv}, I found significant deviations in some locations by comparing the reported coordinates to known locations on a map. While several satellite imagery sources showed similar deviations, it is difficult to determine whether the reported coordinates are erroneous or the map coordinates themselves but regardless, this error is incorporated when I measure the estimated trajectory deviation from ground truth.

\section{Results}

I chose to implement my pose graph optimization in both g2o and Ceres. I compare their performance given my setup in \ref{tab2} over a varying subsets of matched features. As previously described, this value determines how many queried panoramas a given feature must be present in to be considered for bundle adjustment. I find that the g2o solver outperforms Ceres for all of the tested feature subsets, although Ceres appears to be much more robust to the varying features, producing similar results over the different subsets. For my fixed nonlinear BA, I find that taking the intersection of all 4 panoramas produces the best results.

my primary metric as previously described is $RMSE(T_{est})$. I tested over a 1km track in Westwood, CA that saw variable traffic density, vehicle speed, and building density and distance. Over the entire track, I achieved an $RMSE(T_{est})=1.529m$ using my fixed pose bundle adjustment.

\begin{figure*}[h]
\centering
\subfigure[SciPy, fixed pose]{\includegraphics[width=.32\textwidth]{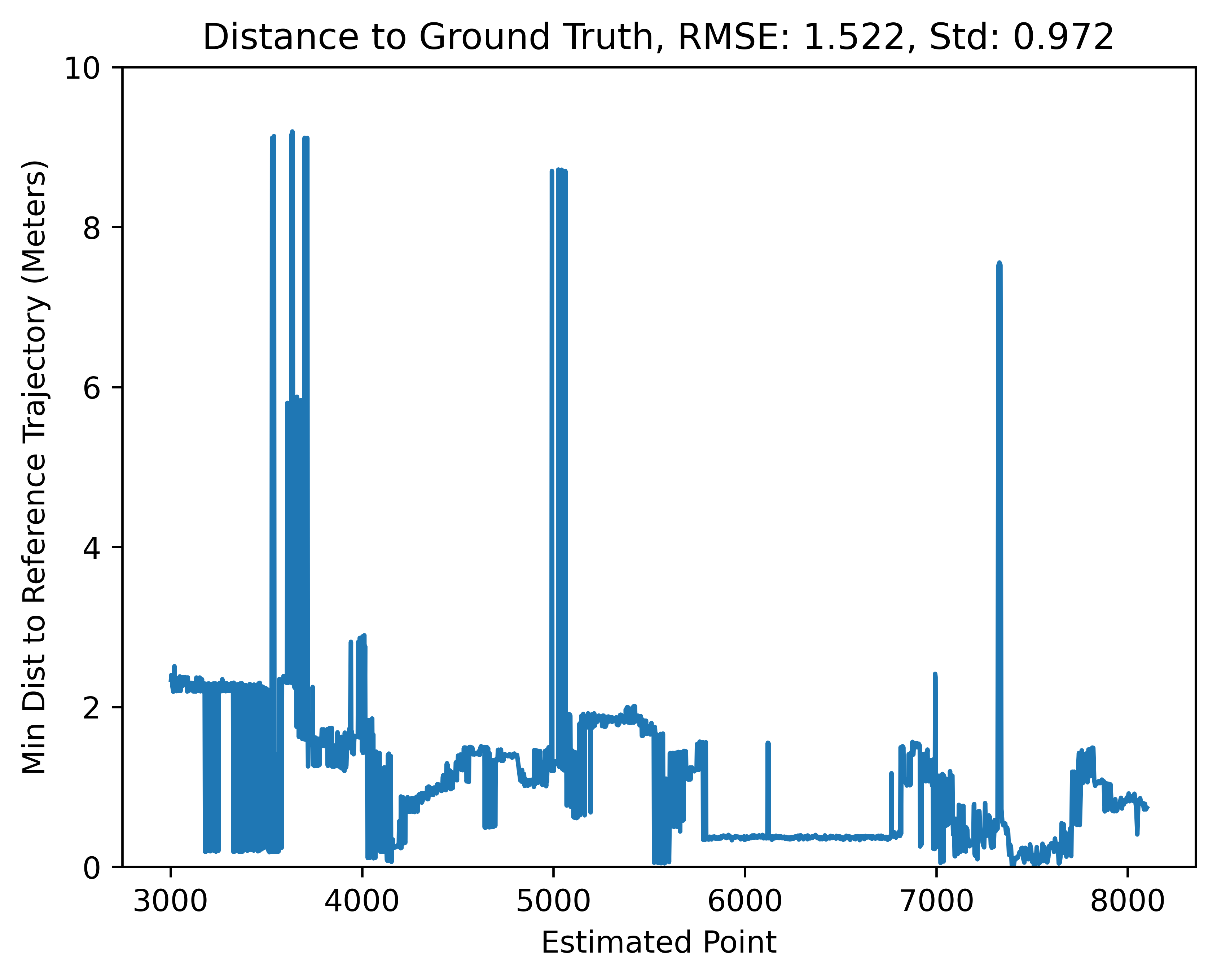} }
\subfigure[g2o, variable pose]{\includegraphics[width=.32\textwidth]{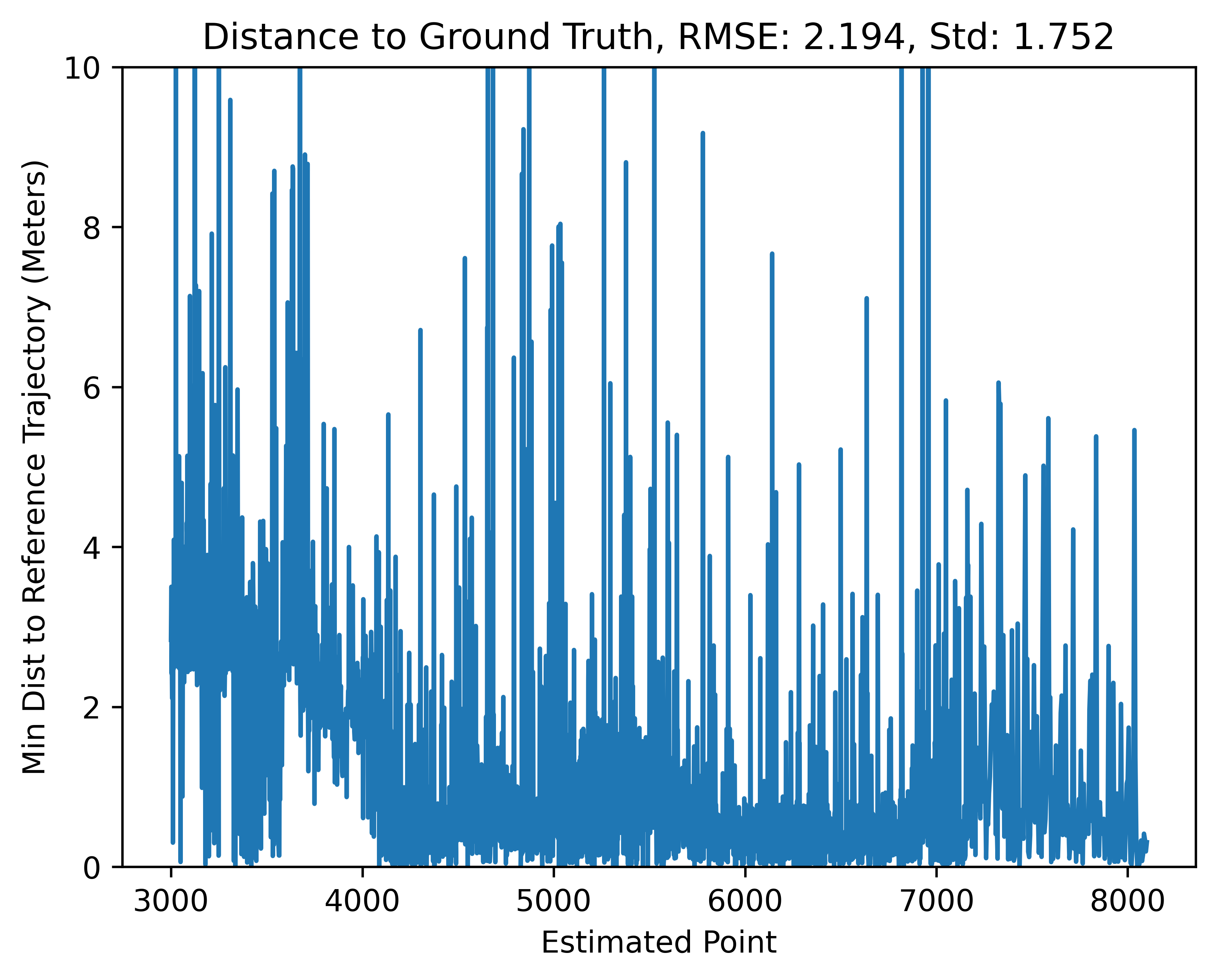} }
\subfigure[Ceres, variable pose]{\includegraphics[width=.32\textwidth]{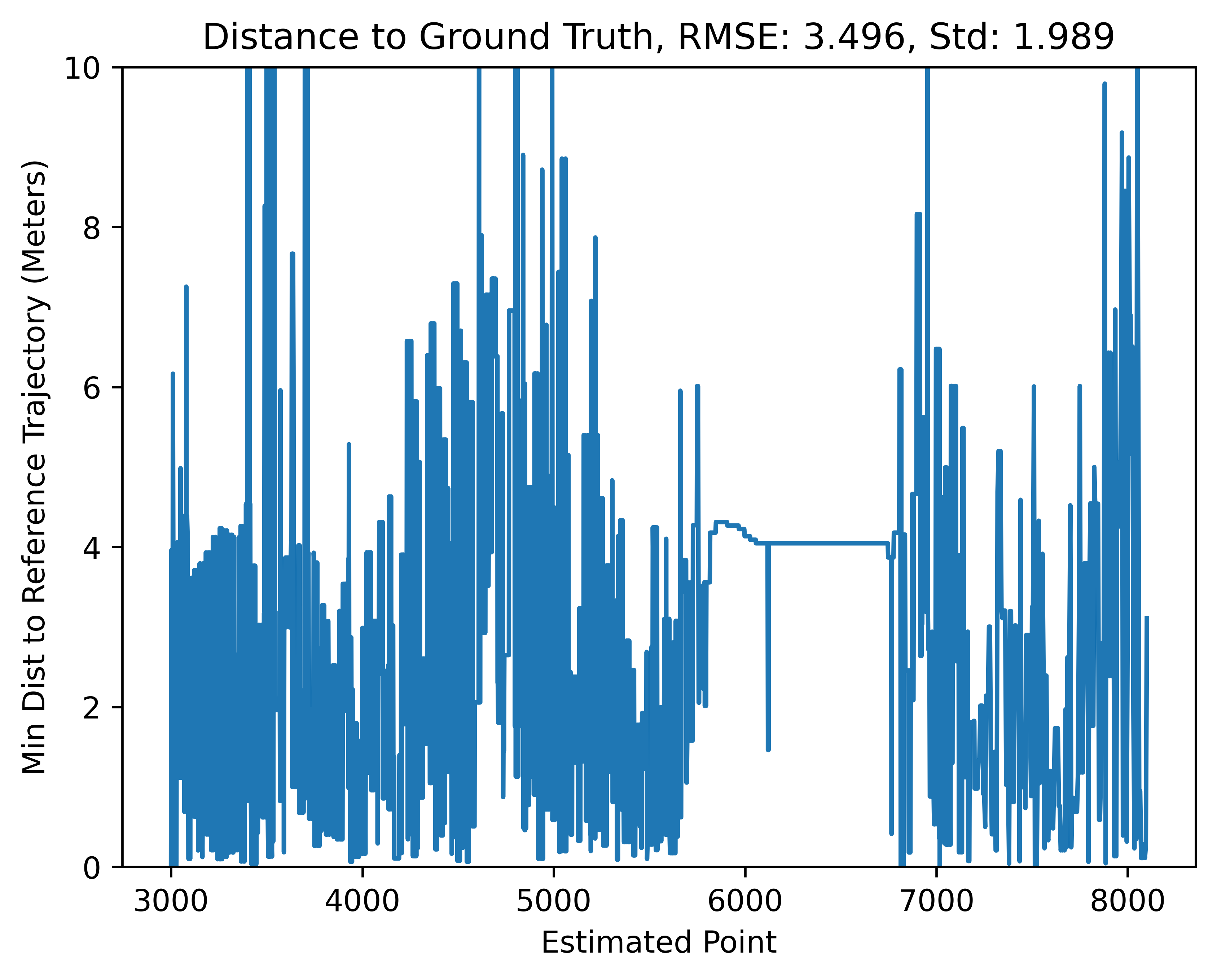} }
\caption{Localization error per frame over a 1 km test track}
\label{fig:rmse}
\end{figure*}

\begin{figure}[h]
\centering
\subfigure[]{\includegraphics[width=0.8\linewidth]{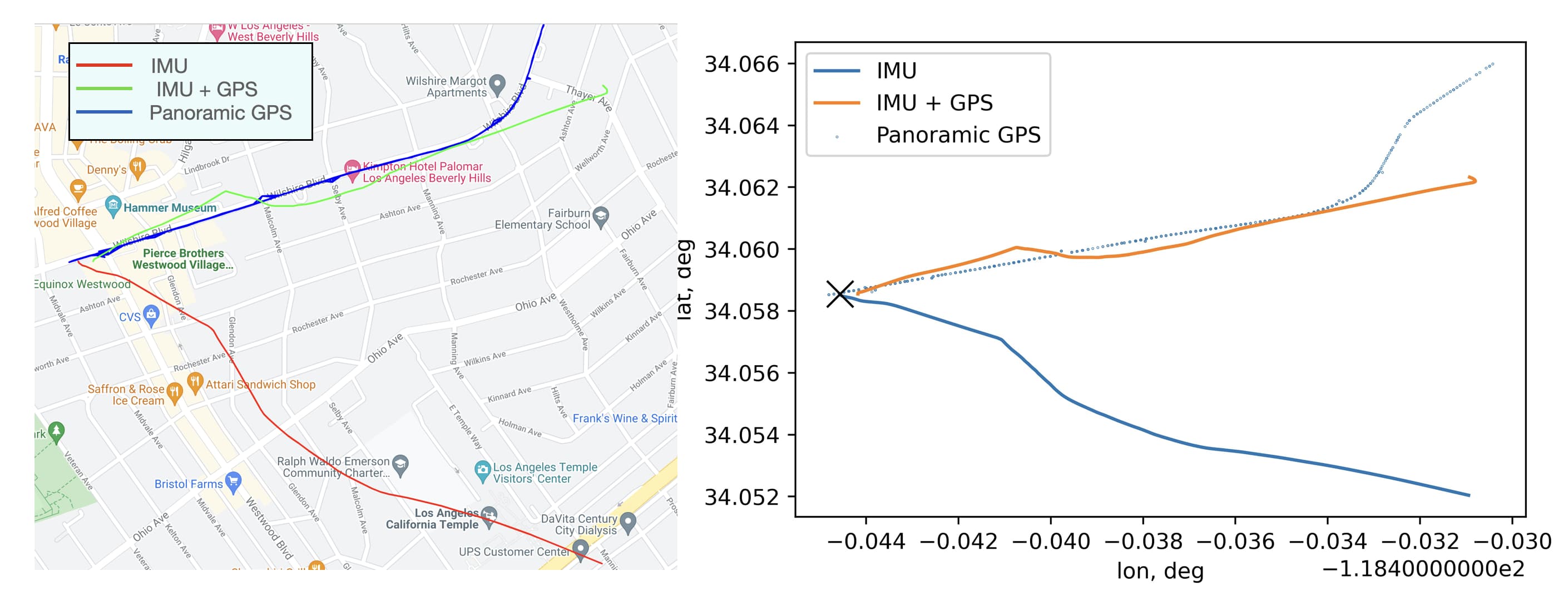}}
\caption{Comparative plot of sensor fusion algorithms}
\label{fig:features1}
\end{figure}

Looking at a plot of $\operatorname{err}(p)$ in Figure \ref{fig:rmse} I see significant variability in the results that appear to be due to the varying environmental conditions. I see frames 5800-6800 have a low error variance which corresponds to a period in which the vehicle is stopped at a light. 

\begin{table}[htbp]
\caption{RMSE and Std Dev over a 100m segment}
\begin{center}
\begin{tabular}{|c|c|c|c|c|}
\hline
\textbf{Min Common Panoramas} & 1 & 2 & 3 & 4 \\
\hline
\textbf{g2o RMSE} & 3.015 &  2.092 & 2.631 & 2.26\\
\hline
\textbf{Ceres RMSE} & 3.596 & 3.596 & 3.596 & 3.596\\
\hline
\textbf{SciPy RMSE} & 2.858 & 2.042 & 1.893 & 1.729\\
\hline
\textbf{g2o Std Dev} & 2.505 & 1.718 & 2.22 & 1.842\\
\hline
\textbf{Ceres Std Dev} & 2.287 & 2.287 & 2.287 & 2.287\\
\hline
\textbf{SciPy Std Dev} & 2.109 & 3.005 & 1.845 & 2.803\\
\hline
\end{tabular}
\label{tab2}
\end{center}
\end{table}

A plot of the trajectory generated by IMU dead reckoning, panoramic GPS and INS + GPS integration is shown in Figure 10. From the plot I can see a significant bias in the gyroscope and accelerometer sensor that causes a drift in the trajectory. The INS + GPS trajectory near the panoramic GPS trajectory but at instances show vast deviation from the road when plotted and analysed visually. There is a scope of improvement in the trajectory generation by filtering out erroneous IMU data, initial calibration of the INS system and better filter tuning based on sensor characteristics. 

\section{Future Work}
There are several aspects of my system that I believe may be significantly improved upon. One area which I did not focus on in this paper is implementing my pipeline directly on a mobile phone running in real-time, however many of the components of my pipeline are already present in many mobile devices. Visual-Inertial Odometry and Extended Kalman Filters are, for example, present on most iOS devices with ARKit and run in real-time. Feature extraction and matching has been demonstrated to run in real-time at 30fps on mobile phones \cite{siftmobile} and newer implementations of SIFT and nearest-neighbor matching on the GPU has shown these both of these steps can be performed in less than 4ms combined \cite{siftgpu}. Performing real-time local bundle adjustment is more challenging, however although implementations have been demonstrated to run in 500ms on a mobile phone \cite{mobileba} and 230ms on a GPU \cite{gpuba}. Finally, while I found VLDs to perform robust outlier rejection for my pipeline, I am unaware of any work to accelerate K-VLD filtering or perform it on a mobile device. Future work may examine the performance of K-VLD filtering on mobile devices and the potential to perform these computations on a remote server and transmit the final results to the client mobile device.

Another future continuation of this work involves combining the odometry estimates of the position from monocular camera and AR poses along with the positional estimate with Google Street View and IMU to have a robust end to end localization. This involves a Visual Inertial Odometry System that combines GPS, IMU and VO data using a Kalman Filter. For additional reading please refer to \cite{7} where the authors have implemented Loosely Coupled Error State Kalman Filter algorithms for autonomous ground vehicles tested on online dataset and \cite{8} where the authors have developed an Multi-state Constrained Extended Kalman Filter.

\section{Conclusion}
my initial motivation for this project was that the sensor suite typically required for autonomous navigation is expensive and inaccessible for the vast majority of the population with motor vehicles. To address this issue, I built a system for accurate localization that is not gated by cost and would be easily accessible to anyone with a car. Using only a monocular camera, IMU, and GPS from a smartphone, along with panoramas retrieved from Google Street View, my method is able to compute a localization and trajectories with visual inertial odometry. Importantly, my method is able to localize with lane level accuracy, which is better than GPS alone, especially in dense city environments with large buildings or structures that obstruct RF signals. Additionally, with advancement in smartphone sensor technology and general smartphone availability, my method will only become more accurate and accessible.

\vspace{12pt}


\begin{thebibliography}{00}
\bibitem{sensorcost} S. LeVine, “What it really costs to turn a car into a self-driving vehicle,” Quartz, 05-Mar-2017. [Online]. Available: https://qz.com/924212/what-it-really-costs-to-turn-a-car-into-a-self-driving-vehicle/. 
\bibitem{sift} D. G. Lowe, “Distinctive image features from scale-invariant key points,” International Journal of Computer Vision, vol. 60, no. 2, pp. 91–110, 2004. 
\bibitem{flann} M. Muja and D. G. Lowe, “Fast approximate nearest neighbors with automatic algorithm configuration,” Proceedings of the Fourth International Conference on Computer Vision Theory and Applications, 2009. 
\bibitem{kvld} Z. Liu and R. Marlet, “Virtual line descriptor and semi-local graph matching method for reliable feature correspondence,” Procedings of the British Machine Vision Conference 2012, 2012. 
\bibitem{g2o}{R. Kuemmerle, G. Grisetti, H. Strasdat, K. Konolige, and W. Burgard,
“g2o: A general framework for graph optimization,” in Proc. IEEE Int.
Conf. Robot. Autom., Shanghai, China, May 2011}
\bibitem{ceres}{ S. Agarwal and K. Mierle. Ceres Solver: Tutorial \& Reference. Google Inc}
\bibitem{gsv}{Google Street View: Capturing the world at street level}
\bibitem{msthesis} {Li Yu. Absolute Localization by Mono-camera for a Vehicle in Urban Area using Street View. Auto-
matic. Université Paris sciences et lettres, 2018.}
\bibitem{metric}{P. Agarwal, W. Burgard, and L. Spinello, “Metric localization using
google street view,” in Intelligent Robots and Systems (IROS), 2015
IEEE/RSJ International Conference on. IEEE, 2015, pp. 3111–3118}
\bibitem{siftgpu}{M. Björkman, N. Bergström and D. Kragic, "Detecting, segmenting and tracking unknown objects using multi-label MRF inference", CVIU, 118, pp. 111-127, January 2014.}
\bibitem{gpuba}{Fixstars, "A CUDA implementation of Bundle Adjustment", 2021. Available: https://github.com/fixstars/cuda-bundle-adjustment}
\bibitem{mobileba}{P. O. Fasogbon, “Depth from Small Motion using Rank-1 Initialization”, arXiv [cs.CV]. 2019.}
\bibitem{siftmobile}{G. Hall, "GPU accelerated feature algorithms for mobile devices", March 2014}
\bibitem{1} Rosten, Edward, and Tom Drummond. "Machine learning for high-speed corner detection." European conference on computer vision. Springer, Berlin, Heidelberg, 2006.
\bibitem{2} Rosten, Edward, and Tom Drummond. "Fusing points and lines for high performance tracking." Tenth IEEE International Conference on Computer Vision (ICCV'05) Volume 1. Vol. 2. Ieee, 2005.
\bibitem{3} Shi, Jianbo. "Good features to track." 1994 Proceedings of IEEE conference on computer vision and pattern recognition. IEEE, 1994.
\bibitem{4} Nistér, David. "An efficient solution to the five-point relative pose problem." IEEE transactions on pattern analysis and machine intelligence 26.6 (2004): 756-770.
\bibitem{5} Savage, Paul G. "Strapdown inertial navigation integration algorithm design part 1: Attitude algorithms." Journal of guidance, control, and dynamics 21.1 (1998): 19-28.
\bibitem{6} Savage, Paul G. "Strapdown inertial navigation integration algorithm design part 2: Velocity and position algorithms." Journal of Guidance, Control, and dynamics 21.2 (1998): 208-221.
\bibitem{9} Thrun, Sebastian. "Probabilistic robotics." Communications of the ACM 45.3 (2002): 52-57.
\bibitem{10} pyINS, https://pyins.readthedocs.io/en/latest/index.html
\bibitem{7} Burusa, Akshay Kumar. "Visual-Inertial Odometry for Autonomous Ground Vehicles." (2017).
\bibitem{8} Mourikis, Anastasios I., and Stergios I. Roumeliotis. "A multi-state constraint Kalman filter for vision-aided inertial navigation." Proceedings 2007 IEEE International Conference on Robotics and Automation. IEEE, 2007.
\bibitem{9} "sfm — openMVG library", Openmvg.readthedocs.io, 2021. [Online]. https://openmvg.readthedocs.io/en/latest/openMVG/sfm/sfm/. [Accessed: 01- Dec- 2021].
\bibitem{mvg}{P. Moulon, P. Monasse, R. Perrot, en R. Marlet, “OpenMVG: Open multiple view geometry”, in International Workshop on Reproducible Research in Pattern Recognition, 2016, bll 60–74.}

\end{thebibliography}
\end{document}